\documentclass[11pt,a4paper]{article}
\usepackage{acl2021}
\usepackage{times}
\usepackage{latexsym}

\usepackage{microtype}

\usepackage{multirow}
\usepackage{latexsym}
\usepackage{booktabs}
\usepackage{courier}
\usepackage{amsmath}
\usepackage{subfig}
\usepackage{placeins}
\usepackage{tikz}
\usepackage{hyperref}
\usepackage{bbm}

\DeclareMathOperator*{\argmax}{argmax}

\begin{document}

\title{Case-Based Abductive Natural Language Inference}

\author{Marco Valentino, Mokanarangan Thayaparan, Andr\'e Freitas\\
Idiap Research Institute, Switzerland \\ University of Manchester, United Kingdom\\ \tt marco.valentino@idiap.ch\\}
\maketitle

\begin{abstract}
Most of the contemporary approaches for multi-hop Natural Language Inference (NLI) construct explanations considering each test case in isolation. However, this paradigm is known to suffer from semantic drift, a phenomenon that causes the construction of spurious explanations leading to wrong conclusions.
In contrast, this paper proposes an abductive framework for multi-hop NLI exploring the retrieve-reuse-refine paradigm in Case-Based Reasoning (CBR). Specifically, we present \emph{Case-Based Abductive Natural Language Inference (CB-ANLI)}, a model that addresses unseen inference problems by analogical transfer of prior explanations from similar examples. 
We empirically evaluate the abductive framework on commonsense and scientific question answering tasks, demonstrating that CB-ANLI can be effectively integrated with sparse and dense pre-trained encoders to improve multi-hop inference, or adopted as an evidence retriever for Transformers. Moreover, an empirical analysis of semantic drift reveals that the CBR paradigm boosts the quality of the most challenging explanations, a feature that has a direct impact on robustness and accuracy in downstream inference tasks. 

\end{abstract}

\section{Introduction}

Multi-hop inference is the task of composing two or more pieces of evidence from external knowledge resources to address a particular reasoning problem \cite{thayaparan2020survey}. In the context of Natural Language Inference (NLI), this task is often used to develop and evaluate explanation-based systems, capable of performing transparent multi-step reasoning with natural language ~\cite{wiegreffe2021teach,jansen2018WorldTree,camburu2018snli}. While multi-hop inference has been largely explored for extractive problems such as open-domain question answering \cite{yang2018hotpotqa}, increasing attention is being dedicated to the abstractive setting, where the models are required to compose long chains of facts expressing abstract commonsense and scientific knowledge \cite{clark2018think,valentino2022hybrid}.
\begin{figure}[t]
\centering
\includegraphics[width=0.9\columnwidth]{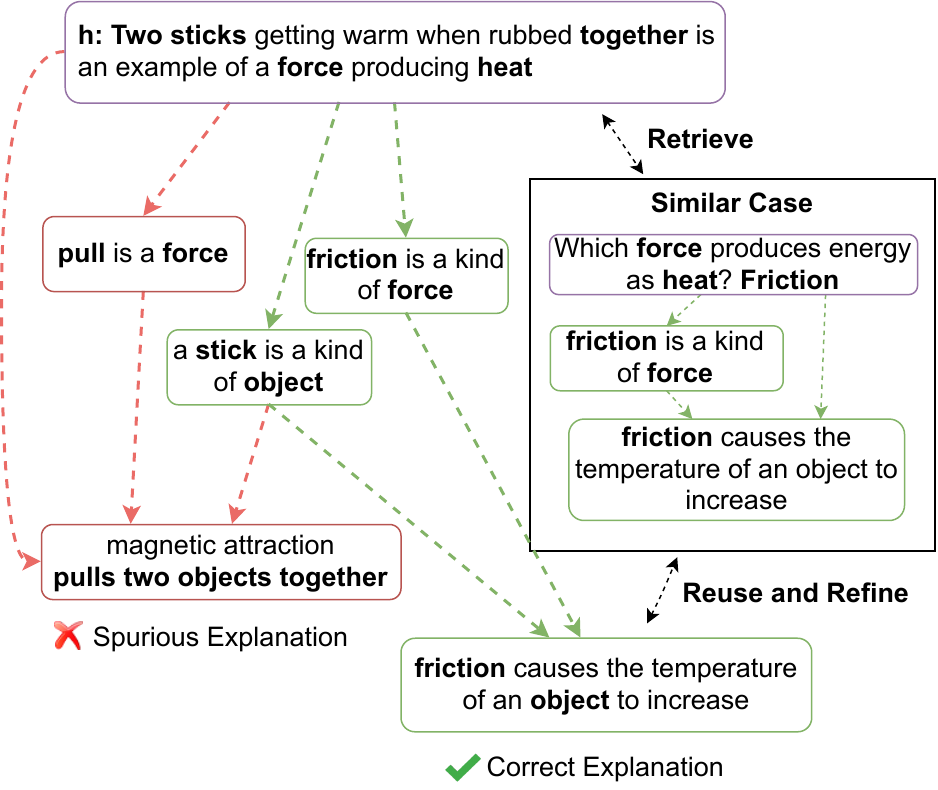}
\caption{Performing multi-hop inference considering each case in isolation can lead to the construction of spurious explanations. In contrast, we propose the adoption of a \emph{Case-Based Reasoning (CBR)} paradigm where the construction of new explanations is constrained by previously solved examples.}
\label{fig:example}
\end{figure}

In this setting, multi-hop inference is often framed as an Abductive Natural Language Inference (ANLI) problem \cite{bhagavatula2019abductive}, where, for a given set of alternative hypotheses $H = \{h_1, h_2, \ldots, h_n\}$, the goal is to construct an explanation for each $h_i \in H$ and select the hypothesis supported by the best explanation \cite{thayaparan-etal-2021-explainable}. 
Existing approaches address abductive inference considering each test hypothesis in isolation, employing iterative and path-based methods \cite{kundu2019exploiting,yadav2019quick} or explicit constraints to guide the generation of a plausible explanation graph supporting the correct answer \cite{khashabi2016question,khot2017answering}.

However, this paradigm poses several challenges in the abstractive setting as: 
\begin{itemize}
    \item The structure of the explanation is not evident from the decomposition of the hypothesis, that is, the type of facts required for the inference cannot be derived from the surface form of the reasoning problem;
    \item Core explanatory facts tend to be abstract, sharing a low number of terms with the hypothesis, making it hard to correctly estimate their relevance for the inference;
    \item Background knowledge sources contain a large amount of irrelevant facts overlapping with the hypothesis, a feature that can lead to the generation of spurious explanations.
\end{itemize}

Consequently, existing approaches often suffer from a phenomenon known as \emph{semantic drift} \cite{khashabi2019capabilities} -- i.e., the tendency of composing incorrect reasoning chains leading to wrong conclusions as the number of required inference steps increases. The example in Figure \ref{fig:example} illustrates some of these challenges.

In contrast with the dominant paradigm, we propose to integrate Abductive Natural Language Inference in a Case-Based Reasoning (CBR) framework \cite{schank2014inside,das2021case}. CBR systems operate under the hypothesis that similar problems require similar solutions, addressing new cases via analogical transfer from previous cases solved in the past. Specifically, the Case-Based Reasoning framework employs a retrieve-reuse-refine paradigm to model inference over unseen problems \cite{ schank2013explanation,de2005retrieval}.
In the context of multi-hop inference, we hypothesise that the adoption of a Case-Based Reasoning framework can help tackle some of the challenges involved in the abstractive setting since:
\begin{itemize}
    \item Similar natural language hypotheses tend to require similar explanations;
    \item Abstract facts tend to express general explanatory knowledge about underlying regularities, being frequently reused to explain a large variety of phenomena;
    \item Prior solutions can explicitly help constraint the search space, reducing the risk of composing spurious inference chains.
\end{itemize}
\begin{figure*}[t]
\centering
\includegraphics[width=0.95\textwidth]{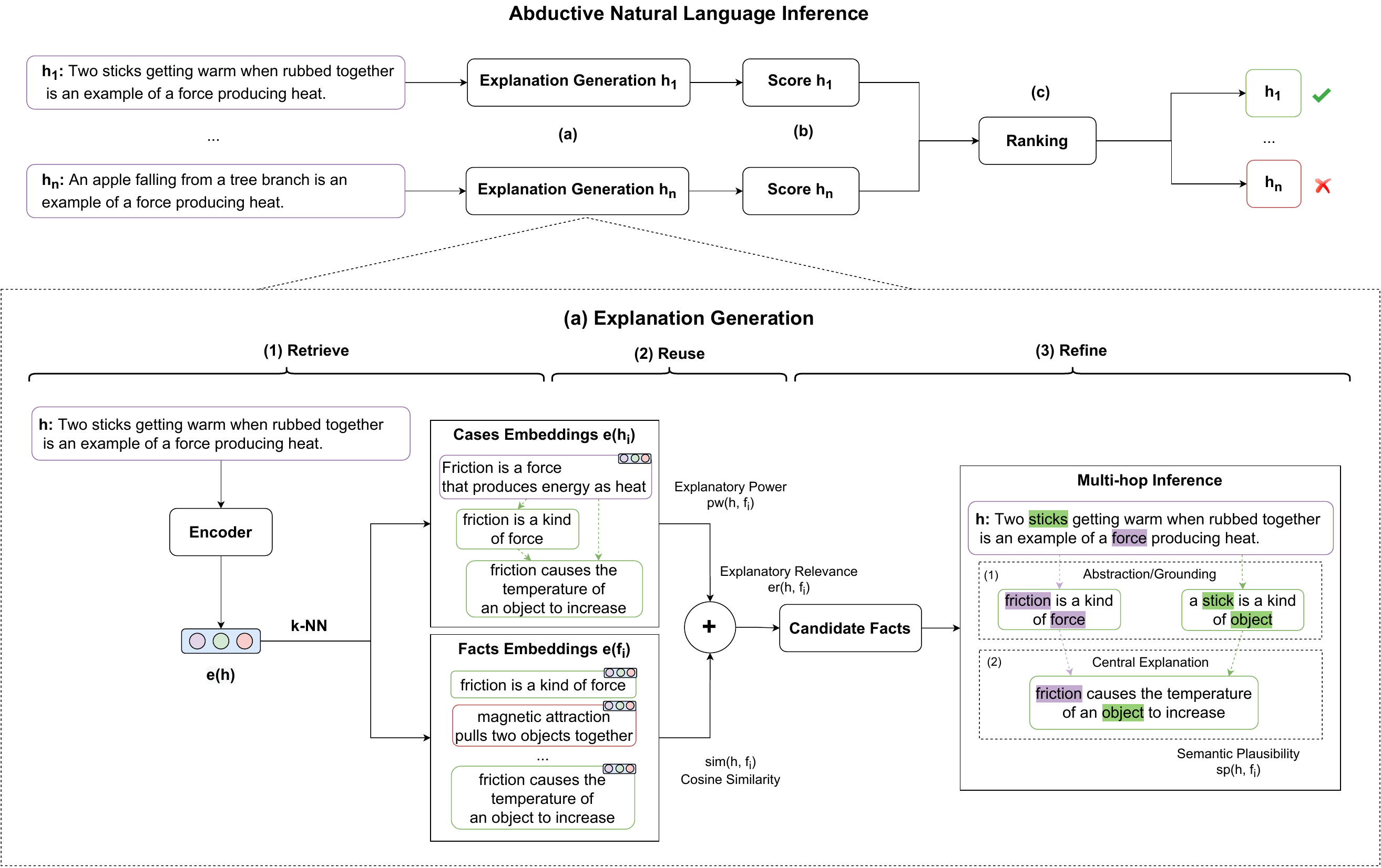}
\caption{Overview of the proposed CBR framework. We adopt a retrieve-reuse-refine paradigm to construct and score explanations for a set of mutually exclusive hypotheses (a) and address NLI tasks via abductive inference by selecting the hypothesis supported by the best explanation.}
\label{fig:approach}
\end{figure*}

To investigate these hypotheses, we present a \emph{Case-Based Abductive NLI (CB-ANLI)} model  that retrieves and adapts natural language explanations from training examples to construct new explanations for unseen cases and address downstream inference problems.
Specifically, this paper provides the following main contributions:
\begin{itemize}
    \item To the best of our knowledge, we are the first to propose an end-to-end case-based abdutive framework for multi-hop and explanation-based NLI;
    \item We empirically demonstrate the efficacy of the CB-ANLI on commonsense and scientific reasoning tasks, showing that the proposed model can be effectively integrated with different sentence encoders and downstream Transformers, achieving strong performance when compared to existing multi-hop and explainable approaches;
    \item We investigate the impact of the retrieve-reuse-refine paradigm on semantic drift, and how this affects accuracy and robustness of the model. Our results show that the case-based framework boosts the quality of the explanations for the most challenging problems, resulting in improved downstream inference performance.
\end{itemize}

\section{Case-based Abductive NLI}
\label{sec:approach}

For a given set of alternative natural language hypotheses $H = \{h_1, h_2, \ldots, h_n\}$, the goal of Abductive NLI is to construct an explanation for each $h_i \in H$ and select the hypothesis supported by the best explanation. Given an hypothesis $h_i$ (e.g., \emph{``Two sticks getting warm when rubbed together is an example of a force producing heat''}), we construct an explanation justifying $h_i$ by extracting and composing inference chains between multiple explanatory facts retrieved from an external corpus. 

To generate and score an explanation for $h_i$, we adopt a Case-Based Reasoning (CBR) paradigm composed of three major phases, retrieve-reuse-refine, which can be summarised as follows (see Fig. \ref{fig:approach}):

\begin{enumerate}
    \item \textbf{Retrieve:}
    In the retrieve phase, we employ a sentence encoding mechanism to search over two distinct embedding spaces. A first embedding space (\emph{Facts Embeddings}) is adopted to retrieve a set of candidate explanatory sentences for the hypothesis.
    A second embedding space (\emph{Cases Embeddings}) is used to retrieve similar cases solved in the past whose explanations can be useful to guide the search for a new solution.
    \item \textbf{Reuse:} In the reuse phase, we condition the relevance of a given fact on the set of explanations retrieved from the most similar cases. Specifically, we reuse previously solved cases to estimate the \emph{explanatory power} of a fact, representing the extent to which a given sentence appears in explanations for past hypotheses.
    \item \textbf{Refine:}
    In this phase, the list of candidate explanatory facts is refined to build the final explanation. We model the construction of an explanation via multi-hop inference between hypothesis and candidate facts, composing abstractive inference chains to estimante the plausibility of the candidate explanatory sentences.
\end{enumerate}

Given a set of alternative hypotheses, we adopt the CBR framework for explanation generation, and subsequently leverage the score assigned to each explanation to address downstream NLI tasks. Additional details on the retrieve-reuse-refine phases are described in the following sections.

\section{Explanation Generation}

\subsection{Retrieve}
\label{sec:analogical_reasoning}

We perform k-NN search over two distinct embedding spaces: (a) an embedding space encoding individual commonsense and scientific facts that can be used to construct new explanations (\emph{Facts Embeddings}); (b) an embedding space of true hypotheses associated with their respective explanations (\emph{Cases Embeddings}). An explanation for a given hypothesis $h_i$ is a composition of facts $E_i = \{f_1, \ldots, f_n\}$ form the Facts Embeddings. 

To perform k-NN search, we employ a sentence encoder $e(\cdot)$. Specifically, we use $e(\cdot)$ to derive a vector for the test hypothesis $h$ and adopt cosine similarity to efficiently score and rank facts and hypotheses in the embedding spaces, retrieving the top-k instances. We perform our experiments using a sparse (BM25 \cite{robertson2009probabilistic}) and a pre-trained dense encoder (Sentence-BERT \cite{reimers2019sentence}) adopting a search index for efficient retrieval (IndexIVFFlat in FAISS \cite{8733051}). We adopt the WorldTree corpus \cite{jansen2018WorldTree} as background knowledge (additional details in Section \ref{sec:emp_eval}).

\subsection{Reuse}

Previous work has shown that explanatory facts expressing underlying regularities tend to create explanatory patterns across similar hypotheses \cite{valentino2022scientific,valentino2021unification,valentino2022hybrid}. Following this line of research, we conjecture that explanations from similar cases can be reused to constraining the search space for unseen hypotheses and improve downstream NLI.  

Specifically, given an unseen hypothesis $h$ and a fact $f_i$, we adopt the explanations retrieved from the top-K similar hypotheses in the \emph{Case Embeddings} to estimate the \emph{explanatory power} of $f_i$:

\begin{equation}
\small
pw(h,f_i) = \sum_{h_k \in kNN(h)}^K {sim(e(h),e(h_k))} \cdot \mathbbm{1}(f_i, h_k)
\label{eq:unification_score}
\end{equation}  

\begin{equation}
\small
    \mathbbm{1}(f_i,h_k) =
        \begin{cases}
            1 & \text{if } f_i \in E_{k}\\
            0 & \text{if } f_i \notin E_{k}
        \end{cases}
\end{equation}
where $kNN(h) = \{h_1, \ldots, h_K\}$ represents the list of k-nearest hypotheses of $h$ retrieved according to the cosine similarity $sim(\cdot)$ between the embeddings $e(h)$ and $e(h_k)$, and $\mathbbm{1}(\cdot)$ is the indicator function verifying if $f_i$ is included in the explanation $E_k$ for the hypothesis $h_k$.
Therefore, for each hypothesis $h_k$ in the set of k-nearest neighbours, the model sums up the quantity $sim(\cdot)$ only if $f_i$ is used to explain $h_k$. Since $sim(e(h),e(h_k))$ represents the similarity between $h$ and $h_k$, the more $f_i$ explains past hypotheses that are similar to $h$ the higher the explanatory power of $f_i$.
To condition the list of candidate explanatory facts on previously solved cases while controlling for relevance with respect to the test hypothesis $h$, we compute the final \emph{explanatory relevance} of each $f_i$ by interpolating the explanatory power with the similarity between the embeddings $e(h)$ and $e(f_i)$:

\begin{equation} 
\small
    er(h, f_i) =  \lambda \cdot sim(e(h),e(f_i)) + (1 - \lambda) \cdot pw(h,f_i)
    \label{eq:model_combination}
\end{equation}
The explanatory relevance score is used to re-rank and filter the list of candidate facts for the subsequent phase.

\subsection{Refine}
\label{sec:abductive_reasoning}

In the refine phase, the model considers the set of candidate facts retrieved in the previous stage to construct the final explanation for $h$.  We model the construction of an explanation through multi-hop inference between hypothesis and candidate facts via the composition of explicit inference chains. To this end, we represent facts and hypothesis as bags of distinct concepts $CP(s_i) = \{cp_1, \ldots, cp_n\}$ (e.g., \emph{``friction is a kind of force''} is represented as the set $\{friction,force\}$, details in the appendix), and connect two generic sentences $s_i$ and $s_j$ by means of shared concept in $CP(s_i) \cap CP(s_j)$. 

To link the hypotheses to potentially abstract explanatory sentences, we construct an explanation graph in different stages, starting with the hypothesis $h$ as the only node. In the first stage, the model extends the graph with the facts that share direct concepts with $h$ and that express taxonomic relations or synonymy. This step can be seen as an abstraction/grounding mechanism aimed at linking the hypothesis to core explanatory facts \cite{jansen2018WorldTree,thayaparan-etal-2021-explainable} (e.g., linking \emph{stick} to \emph{object} and \emph{friction} to \emph{force} in Figure \ref{fig:approach}).

In the second stage, the model extends the graph with all the remaining candidate explanatory facts that share at least one concept with previously added nodes. We consider these facts as \emph{central explanatory nodes}.
After constructing the graph, we estimate the \emph{semantic plausibility} of the central facts $f_i$:
\begin{equation}
\small
sp(h,f_i) = \frac{\sum_{cp_j \in CP(h)} path(cp_j, f_i)}{|CP(h)|}
\label{eq:unification_score}
\end{equation}
where $path(cp_j, f_i)$ is equal to 1 if there exists at least one path in the graph connecting the concept $cp_j$ in the hypothesis to a concept in $f_i$, 0 otherwise.
Therefore, the semantic plausibility of a fact is modelled as the percentage of concepts in the hypothesis $h$ that have at least one path in the graph leading to $f_i$.

\section{Abductive Inference}

To derive the final explanation for a given hypothesis while conditioning on previously solved cases, we sum the \emph{explanatory relevance} computed during the reuse phase with the \emph{semantic plausibility} computed during the refine phase, pruning the graph considering only the top $n$ central explanatory sentences and their linked grounding nodes (Fig. \ref{fig:approach}.3).

Given a set of alternative hypotheses $H = \{h_1, \ldots, h_n\}$, we adopt the model for abductive inference by generating an explanation for each hypothesis and selecting as an answer the one supported by the best explanation. To this end, we assign a score to each hypothesis $h_i$ in $H$ equal to the sum of the scores of the central facts included in the explanation for $h_i$.

\begin{table*}[t]
\centering
\small
\begin{tabular}{p{7cm}|ccccc}
\toprule
\textbf{Model} & \textbf{Overall} & \textbf{Easy} & \textbf{Challenge} & \textbf{Explanation}\\ 
\midrule
\textbf{Sparse Retrieval}\\
\midrule
BM25 ($k$ = 1) \cite{clark2018think} & 41.21 & 44.96 & 32.99 & yes\\
BM25 ($k$ = 2) & 43.62 & 48.54 & 32.73 & \\
BM25 ($k$ = 3) & 45.87 & 50.76 & 35.05 & \\
\midrule
\textbf{Dense Retrieval}\\
\midrule
S-BERT ($k$ = 1) \cite{reimers2019sentence} & 44.91 & 50.99 & 31.44 & yes\\
S-BERT ($k$ = 2) & 45.79 & 51.45 & 33.25 & \\
S-BERT ($k$ = 3) & 44.51 & 49.82 & 32.73 & \\
\midrule
\textbf{Path-based}\\
\midrule
PathNet \cite{kundu2019exploiting} & 41.50 & 43.32 & \textbf{36.42}& yes\\
\midrule
\textbf{Transformers}\\
\midrule
BERT-large \cite{devlin2019bert} & 46.19 & 52.62 & 31.96 & no\\
RoBERTa-large \cite{liu2019roberta}& \textbf{50.20} & \textbf{57.04} & 35.05 & no\\
\midrule
\textbf{Case-based Abductive NLI}\\
\midrule
CB-ANLI BM25 ($n$ = 1) & 52.13 & 56.34  & 42.78  & yes\\
CB-ANLI BM25 ($n$ = 2) & \underline{\textbf{55.17}} & 60.42 &\underline{\textbf{43.56}}  & \\
CB-ANLI BM25 ($n$ = 3) & 52.69 & 58.56 & 39.69  & \\
\midrule
CB-ANLI S-BERT ($n$ = 1) & 54.45 & \underline{\textbf{61.23}} & 39.43  & yes\\
CB-ANLI S-BERT ($n$ = 2) & 52.77 & 59.60 & 37.62  & \\
CB-ANLI S-BERT ($n$ = 3) & 51.64 & 58.67 & 36.08  & \\
\bottomrule
\end{tabular}
\caption{Accuracy on  WorldTree (test-set) for \emph{easy} and \emph{challenge} questions. The parameter $n$ corresponds to the number of central explanatory sentences considered by the models to compute the scores for each hypothesis.}
\label{tab:answer_prediction}
\end{table*}

\section{Empirical Evaluation}
\label{sec:emp_eval}
\paragraph{Experimental Setup.} We evaluate the Case-based Abductive NLI (CB-ANLI) framework on WorldTree \cite{jansen2018WorldTree} and AI2 Reasoning Challenge (ARC) \cite{clark2018think}, two multiple-choice science question answering datasets designed to test abstractive commonsense and scientific inference. 
To perform the experiments, we transform each question-candidate answer pair into a hypothesis following the methodology described in \cite{demszky2018transforming}.

The knowledge bases required for the inference are populated using the WorldTree corpus \cite{jansen2018WorldTree}. The corpus contains a large set of commonsense and scientific facts ($\approx 10K$) that are used to construct explanations for multiple-choice science questions. The explanations include an average of 6 facts (and as many as $\approx$ 20), requiring challenging multi-hop inference to be generated. We store the individual facts for deriving the \emph{Facts Embeddings} and consider the training questions ($\approx$ 1K) and their explanations as the set of previously solved cases (\emph{Cases Embeddings}). For the refine phase, we dynamically extract the concepts in facts and hypotheses using WordNet \cite{miller1995wordnet} with NLTK\footnote{ \url{https://www.nltk.org/_modules/nltk/corpus/reader/wordnet.html}}. Additional details are described in the appendix\footnote{Code available at the following url: \url{https://github.com/ai-systems/case_based_anli}}.

\paragraph{Sentene Encoders.}
We evaluate CB-ANLI using sparse and dense sentence encoders without additional training. The sparse version adopts BM25 vectors \cite{robertson2009probabilistic}, while the dense version employs Sentence-BERT (large)  \cite{reimers2019sentence,thakur-2020-AugSBERT}.

\subsection{WorldTree}

In this section, we present the results achieved on the WorldTree test-set (1247 questions). We report the accuracy of the case-based framework with different numbers $n$ of central facts in the explanations.
We compare the proposed framework against different categories of approaches: \emph{Retrieval Solvers}, \emph{Path-based Solvers}, and \emph{Transformers}. The results in terms of question answering accuracy are reported in Table \ref{tab:answer_prediction}.

\paragraph{Retrieval Solvers.}  We employ stand-alone BM25 and Sentence-BERT (large) as sparse and dense retrieval solvers \cite{clark2018think}. Given an hypothesis $h$, the solvers retrieve the top $k$ relevant facts for $h$ using cosine similarity. The cosine similarity scores are then summed up to determine the best hypothesis. These baselines use the same encoders adopted by our model. However, we observe that CB-ANLI is able to outperform both sparse and dense retrieval models by up to $\approx 10\%$ accuracy, demonstrating the decisive role of the proposed case-based paradigm.

\paragraph{Path-based Solvers.}
We consider PathNet \cite{kundu2019exploiting} as a multi-hop inference baseline. This model constructs inference paths connecting question and candidate answer, and subsequently scores them through a neural encoder to derive the correct answer. We reproduce PathNet using the source code available online\footnote{\url{https://github.com/allenai/PathNet}}. Contrary to CB-ANLI, PathNet does not adopt a Case-Based Reasoning framework to construct the explanations, considering each test hypothesis in isolation. We observe that CB-ANLI can significantly outperform PathNet with up to $\approx 13\%$ improvement overall and $\approx 7\%$ on challenge questions.

\paragraph{Transformers.} We compare CB-ANLI against a BERT large \cite{devlin2019bert} and a RoBERTa large \cite{liu2019roberta} baseline fine-tuned on the multiple-choice question answering task. We observe that on WorldTree the proposed approach is competitive with both RoBERTa and BERT (up to $\approx 5\%$ and $\approx 9\%$ improvement respectively).

\paragraph{Transformers with Explanations}

\begin{table}[t]
\centering
\small
\begin{tabular}{p{3.8cm}ccc}
\toprule
\textbf{RoBERTa + Retriever} & \textbf{Over.} & \textbf{Easy} & \textbf{Chal.} \\ 
\midrule
None & 50.20 &  57.04 & 35.05\\\midrule
BM25 ($k$ = 1) & 57.06 & 60.88 & 48.57\\
BM25 ($k$ = 2 ) & 61.07 & 66.82 & 48.32\\
BM25 ($k$ = 3) & 61.23 & 65.54 & 51.12\\
\midrule
S-BERT ($k$ = 1) &  55.85 & 61.46 & 43.41\\
S-BERT ($k$ = 2) & 60.91 & 66.82 & 47.80\\
S-BERT ($k$ = 3) & 56.96 & 62.04 & 45.73 \\
\midrule
\midrule
CB-ANLI BM25 ($n$ = 1) & 61.71 & 66.82 & 50.38\\
CB-ANLI BM25 ($n$ = 2) & \textbf{63.48} & \textbf{69.38} & 50.38 \\
CB-ANLI BM25 ($n$ = 3) & 62.43 & 67.77 & 50.63\\
\midrule
CB-ANLI S-BERT ($n$ = 1) & 59.99 & 65.54 & 47.45 \\
CB-ANLI S-BERT ($n$ = 2) & 63.32 & 67.98 & \textbf{52.97} \\
CB-ANLI S-BERT ($n$ = 3) & 62.27 & 67.63 & 50.38 \\
\bottomrule
\end{tabular}
\caption{Results for RoBERTa large fine-tuned on WorldTree and augmented with different explanation retrieval models.}
\label{tab:answer_prediction_transformers}
\end{table}

\label{sec:arc_challenge}
\begin{table}[t]
\centering
\small
\begin{tabular}{p{5.5cm}c}
\toprule
\textbf{Explainable Models} & \textbf{Accuracy} \\ 
\midrule
TupleInf \cite{khot2017answering} & 23.83 \\
TableILP \cite{khashabi2016question} & 26.97 \\
DGEM \cite{clark2018think} & 27.11 \\
KG$^2$ \cite{zhang2018kg} & 31.70 \\
Unsupervised AHE \cite{yadav2019alignment} & 33.87 \\
Supervised AHE \cite{yadav2019alignment} & 34.47 \\
ET-RR \cite{ni2019learning} & 36.61 \\
ExplanationLP \cite{thayaparan-etal-2021-explainable} & 40.21\\
AutoROCC \cite{yadav2019quick} & 41.24\\
Attentive Ranker \cite{pirtoaca2019answering} & \textbf{44.72}\\
\midrule
\textbf{Case-based Abductive NLI}\\
\midrule
CB-ANLI BM25 ($n$ = 1) & 33.45\\
CB-ANLI BM25 ($n$ = 2) & 34.39\\
CB-ANLI BM25 ($n$ = 3) & 33.79\\
\midrule
CB-ANLI S-BERT ($n$ = 1) & \textbf{36.77}\\ 
CB-ANLI S-BERT ($n$ = 2)& 35.75 \\
CB-ANLI S-BERT ($n$ = 3) & 34.30 \\
\midrule
\midrule
CB-ANLI S-BERT ($n$ = 1) + RoBERTa & 44.02\\
CB-ANLI S-BERT ($n$ = 2) + RoBERTa & \underline{\textbf{47.86}} \\
CB-ANLI S-BERT ($n$ = 3) + RoBERTa & 42.40 \\
\bottomrule
\end{tabular}
\caption{Performance on the AI2 Reasoning Challenge (ARC). We compare CB-ANLI with published explainable approaches that are fine-tuned only on ARC. }
\label{tab:answer_prediction_arc}
\end{table}

We evaluate CB-ANLI as an evidence retrieval model by combining it with downstream Transformers. To perform this experiment, we augment the input of RoBERTa large with the explanations constructed for each hypothesis, and fine-tune the model to maximise the score for the correct answer.
Table \ref{tab:answer_prediction_transformers} reports the accuracy achieved with RoBERTa large when adopting CB-ANLI and stand-alone baselines as evidence retrievers. In general, we observe that evidence retrieval plays an important role for improving the performance of RoBERTa, and that the use of CB-ANLI can generate useful explanations for inference in combination with downstream language models. 

\begin{figure*}[t]
\centering
\subfloat[]{\includegraphics[width=\columnwidth]{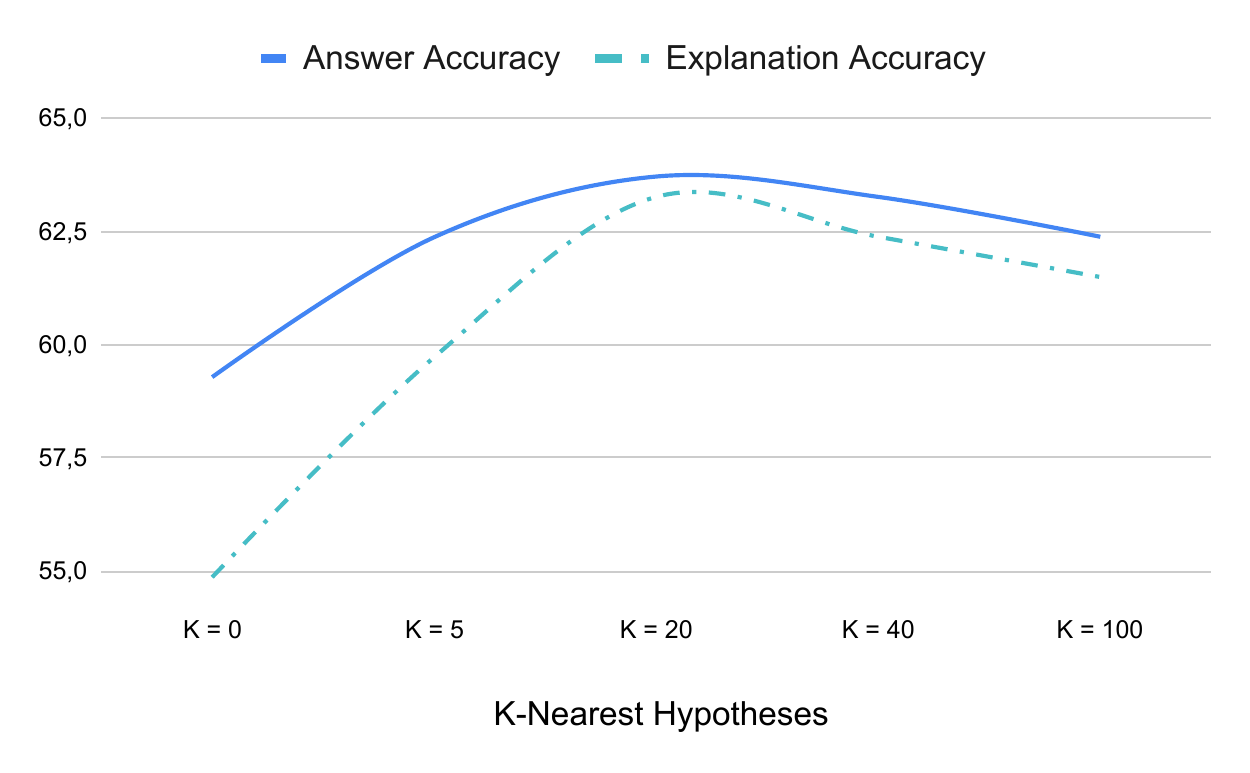}}
\subfloat[]{\includegraphics[width=\columnwidth]{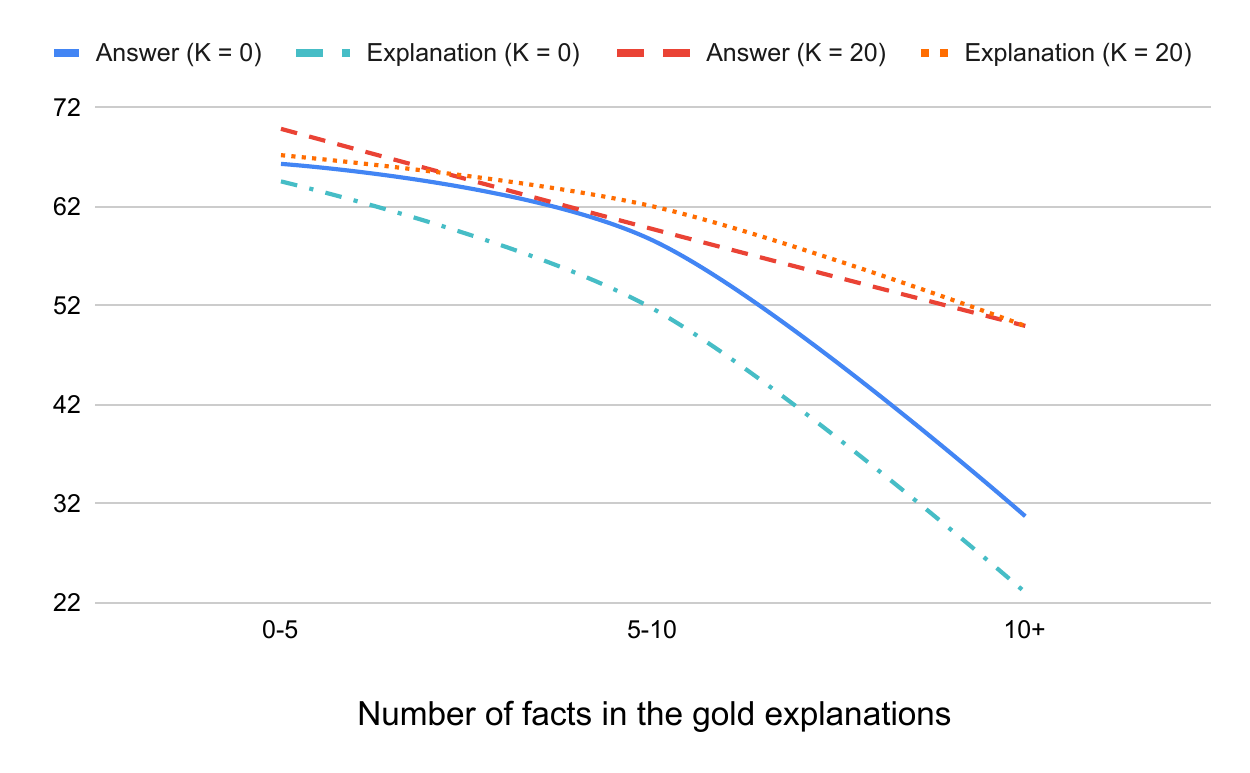}}
\caption{Impact of the case-based framework on semantic drift. $K$ represents the number of similar cases considered for computing the explanatory power (Worldree dev-set).}
\label{fig:explanation_eval}
\end{figure*}

\subsection{ARC Challenge}

To evaluate the generalisation of CB-ANLI on a broader set of challenge questions, we run additional experiments on the AI2 Reasoning Challenge (ARC) \cite{clark2018think}. 
Here, we keep the same configuration of hyperparameters. 
Table \ref{tab:answer_prediction_arc} reports the results achieved on the test-set (1172 challenge questions).

We observe that CB-ANLI with Sentence-BERT can generalise better on ARC. We attribute these results to the ability of Sentence-BERT to go beyond lexical overlaps, supporting generalisation on new hypotheses with different surface forms. To show the impact of evidence retrieval on ARC, we fine-tune RoBERTa with the explanations constructed by the Sentence-BERT version.

For a fair comparison, we compare CB-ANLI against published explainable approaches that are fine-tuned only on ARC, without additional pre-training on related datasets (e.g. OpenBookQA \cite{mihaylov2018can}, RACE \cite{lai2017race}).
The results show that CB-ANLI S-BERT is third in the ranking, outperforming explanation-based systems based on Integer Linear Programming (ILP) \cite{khot2017answering,khashabi2016question} and pre-trained embeddings \cite{yadav2019alignment}. At the same time, CB-ANLI obtains competitive results when compared with most of the fine-tuned neural approaches, including ET-RR \cite{ni2019learning}. 
Moreover, when combined with RoBERTa, CB-ANLI achieves the best results among the considered approaches, improving on AutoROCC  \cite{yadav2019quick} and Attentive Ranker \cite{pirtoaca2019answering}.

\subsection{Ablation Study}

\begin{table}[t]
\centering
\small
\begin{tabular}{p{2.9cm}ccc}
\toprule
\textbf{Paradigm} & \textbf{Overall} & \textbf{Easy} & \textbf{Challenge}\\ 
\midrule
\textbf{CB-ANLI BM25} & & \\
\midrule
Retrieve-Reuse-Refine & \textbf{55.17} & \textbf{60.42} & \textbf{43.56}\\
Retrieve-Reuse & 49.00 & 55.18 & 35.30\\
Retrieve-Refine & 43.46 & 46.57 & 36.60\\
\midrule
\textbf{CB-ANLI S-BERT} & & \\
\midrule
Retrieve-Reuse-Refine & \textbf{54.45} & \textbf{61.23} & \textbf{39.43}\\
Retrieve-Reuse & 47.79 & 53.55 & 35.05\\
Retrieve-Refine & 42.66 & 47.48 & 32.21\\
\bottomrule
\end{tabular}
\caption{  Ablation Study on WorldTree (test-set) by removing the impact of the reuse and refine phases.}
\label{tab:ablation}
\end{table}

\begin{table*}[t]
\centering
\tiny
\begin{tabular}{p{6cm}p{1.5cm}p{6cm}c}
\toprule
\textbf{Test Question} & \textbf{Prediction} & \textbf{Constructed Explanation ($K = 20, n = 1$)} & \textbf{Accurate}\\
\midrule
What force is needed to help stop a child from slipping on ice? (A) gravity, (B) \underline{friction}, (C) electric, (D) magnetic 
& (B) \underline{friction} 
& (1) counter means reduce; stop; resist;
(2) ice is a kind of object;
(3) slipping is a kind of motion;
(4) stop means not move; \textbf{(5) friction acts to counter the motion of two objects when their surfaces are touching}
& \textbf{Y}\\
\midrule
What causes a change in the speed of a moving object? (A) \underline{force}, (B) temperature, (C) change in mass (D) change in location
& (A) \underline{force}
& \textbf{(1) a force continually acting on an object in the same direction that the object is moving can cause that object's speed to increase in a forward motion} & N\\
\midrule
Weather patterns sometimes result in drought. Which activity would be most negatively affected during a drought year? (A) boating, (B) \underline{farming}, (C) hiking, (D) hunting
& (B) \underline{farming}
& (1) affected means changed;
(2) a drought is a kind of slow environmental change; \textbf{(3) farming changes the environment} & N\\
\midrule
Beryl finds a rock and wants to know what kind it is. Which piece of information about the rock will best help her to identify it? (A) The size of the rock, (B) The weight of the rock, (C) The temperature where the rock was found, (D) \underline{The minerals the rock contains}
& (A) The size of the rock
& (1) a property is a kind of information;
(2) size is a kind of property;
(3) knowing the properties of something means knowing information about that something.
\textbf{(4) the properties of something can be used to identify; used to describe that something} & \textbf{Y}\\
\midrule
Jeannie put her soccer ball on the ground on the side of a hill. What force acted on the soccer ball to make it roll down the hill? (A) \underline{gravity}, (B) electricity, (C) friction, (D) magnetism
& (C) friction
& (1) the ground means Earth's surface;
(2) rolling is a kind of motion;
(3) a roll is a kind of movement; \textbf{(4) friction acts to counter the motion of two objects when their surfaces are touching}
& N \\
\bottomrule
\end{tabular}
\caption{ Examples of explanations constructed by CB-ANLI. The  \underline{underlined choices} represent the correct answers. \emph{Accurate} indicates whether the central fact \textbf{(bold)} is part of the gold explanation in the WorldTree corpus.}
\label{tab:qualitative_examples}
\end{table*}



We carried out an ablation study to investigate the impact of the CBR framework on downstream inference performance. To this end, we consider different versions of CB-ANLI by removing the impact of the reuse and refine phase.
For the first, we remove the explanatory power term in equation 3. For the latter, we simply skip the refine phase ignoring the explanation graph construction and the semantic plausibility score to filter the central explanatory facts. The results of the study, reported in Table \ref{tab:ablation}, demonstrate the key role of each phase to achieve the final inference performance.

\subsection{Impact on Semantic Drift}
In this section, we investigate the impact of the CBR paradigm on semantic drift and how this affects the results on downstream reasoning tasks. 
To this end, we measure the performance of CB-ANLI when retrieving a different number $K$ of previously solved hypotheses (notice that when $K = 0$ the model is equivalent to a non-case-based method). 
To evaluate the quality of the generated explanations, we leverage the WorldTree corpus as a gold standard, computing the \emph{explanation accuracy} as the percentage of the best central explanations retrieved by the model that are part of the gold explanations in WorldTree. Since the explanations in the test-set are not publicly available, we perform this analysis on the dev-set.

Figure \ref{fig:explanation_eval} (a) illustrates the change in answer and explanation accuracy on WorldTree with an increasing number $K$ of similar cases. The graph demonstrates that the improvement in answer prediction is associated with improved explanation generation capabilities. Specifically, by conditioning the inference on an increasing number of similar hypotheses, CB-ANLI is able to construct more accurate explanations, a feature that has a direct impact on downstream inference performance in question answering. 

Figure \ref{fig:explanation_eval} (b) shows the accuracy of the model on hypotheses requiring longer explanations when compared to a non-case-based version ($K=0$). In general, a higher number of facts in the gold explanation is associated with a higher probability of semantic drift \cite{jansen2019textgraphs}. The graph confirms a strong relation between explanation accuracy and question answering accuracy, and demonstrates that the improvement obtained through the case-based framework is particularly evident on the most challenging inference problems (10+ facts in the explanations). This results allow us to conclude that the Case-Based Reasoning framework has a key role in alleviating semantic drift during multi-hop inference. 

\subsection{Faithfulness and Error Analysis}
Finally, we present an analysis of the faithfulness of the model, investigating the relation between correct/wrong answer prediction and accurate/inaccurate explanations. Overall, we found that a total of 81.25\% of the correct answers are derived from accurate explanations. This situation is illustrated in the first example in Table \ref{tab:qualitative_examples}. On the other hand, a total of 18.75\% of correct answers are derived from inaccurate explanations (second and third rows in the table). However, as shown in the second example, we observe that CB-ANLI can sometimes find alternative ways of constructing plausible explanations, considered inaccurate only because of a mismatch with the corpus annotation. The example number 4 shows the case in which an accurate explanation is not sufficient to discriminate the correct answer. We found this cases to  occur for a total of 31.71\% of incorrect answers. Finally, the last row describes the situation in which wrong answers are caused by inaccurate or spurious explanations (for a total of 68.29\% of the wrong answers). This analysis demonstrates the interpretability and faithfulness of the framework, showing that its behaviour can be typically traced back to the quality of the generated explanations.



\section{Related Work}

Multi-hop inference for abstractive tasks is challenging as the general structure of the explanations cannot be derived from the surface form of the NLI problem. Previous work has demonstrated that models in this setting are affected by semantic drift -- i.e., the construction of spurious explanations leading to wrong conclusions \cite{fried2015higher,khashabi2019capabilities}.

Existing approaches frame multi-hop inference as the problem of building an optimal graph, conditioned on a set of semantic constraints \cite{khashabi2018question,khot2017answering,jansen2017framing,khashabi2016question,thayaparan-etal-2021-explainable}, or adopting iterative methods, using sparse or dense encoding mechanisms \cite{yadav2019alignment,yadav2019quick,pirtoaca2019answering,kundu2019exploiting}.
Our model is related to previous work that leverages annotated explanations to reduce semantic drift \cite{xie2020WorldTree,jansen2018WorldTree}. However, this line of work is still limited to explanation regeneration \cite{jansen2019textgraphs,cartuyvels2020autoregressive,valentino2021unification,valentino2022hybrid,thayaparan2021textgraphs}, while the applicability of these resources for downstream multi-hop NLI problems is yet to be explored. In this paper, we move a step forward, exploring how the impact of annotated explanations on semantic drift translates in improved downstream performance.

Our approach is related to previous work on Case-Based Reasoning (CBR)~\cite{schank2014inside, schank2013explanation,de2005retrieval}. Similar to the retrieve-reuse-refine paradigm adopted in CBR systems, we employ encoding mechanisms to retrieve explanations for cases solved in the past, and adapt them in the solution of new problems.
Recent work in NLP investigates the use of a similar paradigm via k-NN retrieval on training examples. \citet{Khandelwal2020Generalization,khandelwal2020nearest} adopt k-NN search to retrieve similar training examples and improve pre-trained language models and machine translation without additional training. Similarly, \citet{das2021case,das2020simple} propose a CBR framework for knowledge base reasoning, while \citet{kassner-schutze-2020-bert} reuse similar cases to improve BERT \cite{devlin2019bert} on cloze-style QA. To the best of our knowledge, this is the first application of Case-Based Reasoning for explanation-based multi-hop Natural Language Inference (NLI).

The work presented in this paper is related to hybrid neuro-symbolic approaches for inference with natural language  \cite{liu2020multi,minervini2020differentiable,jiang2019self,chen2019neural,dua2019drop,xu2021exploiting,weber2019nlprolog}. In this context, most of the existing approaches combine neural models with symbolic programs or reasoning modules. For instance, \citet{jiang2019self} propose the adoption of a Neural Module Network~\cite{andreas2016neural} for multi-hop question answering by designing four atomic neural modules (Find, Relocate, Compare, NoOp). \citet{weber2019nlprolog} propose a methodology to perform multi-hop inference using a Prolog  prover via the integration of sentence encoders and a weak unification mechanism. Differently from the methodology discussed in this paper, previous neuro-symbolic approaches have been generally applied to extractive tasks, where the inference steps (and, therefore, the explanation's structure) can be derived from a direct decomposition of the questions \cite{thayaparan2020survey}. 


\section{Limitations}

The adopted model of explanatory power relies on the availability of human-annotated explanations with specific features (e.g., explanatory facts reused across different training instances). However, these resources might not be available in real-world scenarios and are generally costly to develop. Moreover, since the explanatory power model relies on similarity measures and indicator functions, the model's ability to generalise might be sensitive to the incompleteness of the knowledge bases and the availability of representative explanations. We believe these limitations can be potentially alleviated by exploring the role of more abstract sentence representations within the CBR paradigm \cite{bergmann1996role}. 

In the current implementation of CB-ANLI, the refine phase adopts specific assumptions to model the abstraction process required for explanation generation. This process, in fact, is performed by assuming that abstraction at the concept level translates in a correct mapping between hypotheses and central explanatory sentences. However, contextual linguistic elements can still affect the overall meaning of the specific concept being abstracted, inducing the inclusion of spurious links between sentences. While contextual elements are considered during the precedent phases through the use of contextualised embeddings and similar cases, additional work is still required to guarantee the correctness of the abstraction process. 

\section{Conclusion}
This paper presented CB-ANLI, a model that integrates multi-hop and Case-Based Reasoning (CBR) in a unified framework.  We demonstrated the efficacy of the framework in complex abstractive and multi-hop NLI tasks. We believe this work can open new lines of research on hybrid neuro-symbolic models for explanation-based NLI, and plan to investigate the efficacy of the framework on architectures that adopt richer symbolic representations in combination with neural models, further exploring the role of abstraction in Case-Based Reasoning for improving robustness, generalisation, and explainability in NLI.

\section*{Acknowledgements}

The work is partially funded by the SNSF project NeuMath (200021\_204617). We would like to thank the Computational Shared Facility of the University of Manchester for providing the infrastructure to support our experiments. 

\bibliography{coling2022}
\bibliographystyle{acl_natbib}

\appendix

    \section{Hyperparameters}
   
    We adopted the following hyperparameters for CB-ANLI:
    
    \paragraph{CB-ANLI BM25:}
    \begin{enumerate}
        \item $\lambda$ = 0.83
        \item $K = 200$
    \end{enumerate}
    
    \paragraph{CB-ANLI S-BERT:}
    \begin{enumerate}
        \item $\lambda$ = 0.97
        \item $K = 40$
    \end{enumerate}
    For the implementation of Sentence-BERT we adopt the following package \url{https://pypi.org/project/sentence-transformers/} considering the \emph{bert-large-nli-stsb-mean-tokens} model.
    
    \section{Concepts Extraction}
    The concepts in facts and hypotheses are extracted using WordNet with NLTK: \url{https://www.nltk.org/_modules/nltk/corpus/reader/wordnet.html}.
    Specifically, given a sentence, we define a concept as a maximal sequence of words that corresponds to a valid synset in WordNet. This allows us to consider multi-words expressions such as \emph{``living thing''} that frequently occur in the scientific domain.
    
    \section{Transformers Setup}
    For the implementation of the Transformer model, we fine-tuned RoBERTa (\emph{roberta-large}) for binary classification ($bc$) to predict a set of scores  $S = \{s_1,~s_2,~...,~s_n\}$ for each candidate hypothesis in $H = \{h_1,~h_2,~...,~h_n\}$. The model receives as input an hypothesis $h_i$ along with the explanation $E_i$ for $h_i$. The model is optimised via cross-entropy loss to predict 1 for the correct hypothesis and 0 for the alternative hypotheses:
    \begin{equation}
        bc(\textrm{\texttt{[CLS]}}~||~h_i~||~\textrm{\texttt{[SEP]}}~||~E_{i}) = s_i
    ~\label{eq:bert_answer}
    \end{equation}
    The binary classifier is a linear layer operating on the final hidden state encoded in the \texttt{[CLS]} token.
    To answer the question $q$, the module selects the candidate answer $c_a$ associated to the hypothesis with the highest score -- i.e. $a = \argmax_i s_i$.
    The model is implemented using Hugging Face (\url{https://huggingface.com/}) and fine-tuned using 4 Tesla V100 GPUs for 8 epochs in total.
    We adopted the following hyperparameters:
    \begin{itemize}
        \item batch size = 16
        \item learning\_rate = 1e-5
        \item gradient\_accumulation\_steps = 1
        \item weight\_decay = 0.0
        \item adam\_epsilon = 1e-8
        \item warmup\_steps = 0
        \item max\_grad\_norm = 1.0
    \end{itemize}

    \section{Source Code}
    The code adopted in the experiments is available at the following URL: \url{https://github.com/ai-systems/case_based_anli}.
    
    \section{Data}
    The WorldTree corpus adopted in the experiments can be downloaded at the following url: \url{http://cognitiveai.org/dist/worldtree_corpus_textgraphs2019sharedtask_withgraphvis.zip}. The AI2 Reasoning Challenge (ARC) dataset can be downloaded at the following URL: \url{https://allenai.org/data/arc}. For the experiments on ARC, we adopted WorldTree V2 as our background knowledge: \url{http://cognitiveai.org/explanationbank/}

\end{document}